\documentclass[final]{cvpr}

\usepackage{times}
\usepackage{epsfig}
\usepackage{graphicx}
\usepackage{amsmath}
\usepackage{amssymb}
\usepackage{bm} 
\usepackage{siunitx}
\usepackage{makecell}

\DeclareMathOperator*{\argmin}{arg\,min}

\usepackage[pagebackref=true,breaklinks=true,colorlinks,bookmarks=false]{hyperref}

\def\cvprPaperID{****} 
\def\confYear{CVPR 2021}

\newcommand{\jhat}{\boldsymbol{\hat{\jmath}}}

\begin{document}

\title{Physically Plausible Pose Refinement using Fully Differentiable Forces}

\author{Akarsh Kumar \quad Aditya R. Vaidya \quad Alexander G. Huth\\
The University of Texas at Austin\\
{\tt\small \{akarshkumar0101,avaidya\}@utexas.edu, huth@cs.utexas.edu}
}

\maketitle

\begin{abstract}
   All hand-object interaction is controlled by forces that the two bodies exert on each other, but little work has been done in modeling these underlying forces when doing pose and contact estimation from RGB/RGB-D data.
   Given the pose of the hand and object from any pose estimation system, we propose an end-to-end differentiable model that refines pose estimates by learning the forces experienced by the object at each vertex in its mesh.
   By matching the learned net force to an estimate of net force based on finite differences of position, this model is able to find forces that accurately describe the movement of the object, while resolving issues like mesh interpenetration and lack of contact.
   Evaluating on the ContactPose dataset, we show this model successfully corrects poses and finds contact maps that better match the ground truth, despite not using any RGB or depth image data. 
\end{abstract}

\section{Introduction}

Hand-object interaction is a long-studied problem in computer vision due to its numerous applications in fields like robotics, neuroscience, and human-computer interaction~\cite{brahmbhattContactPoseDatasetGrasps2020,hassonLearningJointReconstruction2019,moonInterHand26MDataset2020}.
Particularly in the case of egocentric perception, understanding hand-object interaction has implications in augmented reality (AR) and virtual reality (VR).

When watching a video, humans can deduce when a hand is touching an object by observing their relative motion.
For example, if an object is floating mid-air, one can conclude the hand is holding the object up, since something must be counteracting its gravity.
Our proposed model uses this intuition to reason about hand-object contact forces in object manipulation.
If this model is given noisy pose information and can refine the poses to find physically plausible forces that explain the motion, we claim this correction is accurate relative to ground truth pose and contact.

In this work, we propose a pose refinement system that uses insights from mechanics to reason about the physics of a scene.
This system takes as input an initial guess of hand and object pose, rather than data from a video.
This modularity lends us flexibility in applications, because any pose estimation system can give the initial poses. 
Through end-to-end optimization, our system simultaneously finds plausible forces that explain the motion of the object, while also refining hand and object pose.

\section{Related work}

Previous work either does not use the motion of the hand and object to inform their poses~\cite{brahmbhattContactPoseDatasetGrasps2020, moonInterHand26MDataset2020}, or does not capture these interactions in a differentiable manner~\cite{hampaliHOnnotateMethod3D2020,tzionasCapturingHandsAction2016}.

Even though image-based hand-object pose estimation systems cannot use motion information, they are still taught to prefer ``stable'' configurations, such that the object does not fall past the hand during a physical simulation.
To enforce this constraint, Hasson \etal~\cite{hassonLearningJointReconstruction2019} trained neural networks to bias fingers closer to the object's surface if they increase grasp stability.
While static scenes may be a fair assumption for images, natural object manipulation is never truly static due to the motion of the hand and object.

By using video instead of images, models can use temporal information to improve pose estimation.
One such approach favors temporal consistency by minimizing estimated velocities and accelerations for the hand and object~\cite{hampaliHOnnotateMethod3D2020}.
This approach, however, only assumes low velocity and acceleration and does not consider object contact, which may produce larger changes in momentum.

Tzionas \etal~\cite{tzionasCapturingHandsAction2016} do use physics and hand-object contact to inform their poses during estimation, but they achieve this with an expensive, non-differentiable physics simulator.
This choice complicates their use of a gradient-based optimizer.
Additionally, it requires the simplifying assumption that every frame is a static scene.

HO-3D~\cite{hampaliHOnnotateMethod3D2020} and ContactPose~\cite{brahmbhattContactPoseDatasetGrasps2020} are two recent datasets comprising videos of hands manipulating objects with static grasps.
ContactPose additionally includes infrared images of the object, which provide ground truth hand-object contact maps.
In contrast with hand and object pose estimation, annotating these contact maps requires no human judgement, and is thus free from human perceptual biases.
We used ContactPose to evaluate our physics-based pose refinement system.

\section{Methods}
\label{sec:methods}
Inspired by previous literature~\cite{hampaliHOnnotateMethod3D2020,tzionasCapturingHandsAction2016}, we formulate our refinement problem as an energy minimization problem with an energy function that models the physics underlying the scene.
We assume the hand is the only actuator in this scene, and thus the only forces on the object are the result of hand contact and gravity.

\subsection{Optimization variables}
The object is treated as a rigid triangular mesh, 
while the hand is described with MANO, a parametrized
linear blend skinning model~\cite{romeroEmbodiedHandsModeling2017, hassonLearningJointReconstruction2019}.
For a video with $T$ frames, we represent the given initial pose of the hand and object with degrees of freedom (DoF) $\theta_h\in \mathbb{R}^{T\times 21}$ and $\theta_o\in \mathbb{R}^{T\times 6}$, respectively.
The first six elements of each represent a rigid transformation, with the rotation in axis-angle representation.
The last fifteen elements of the hand pose represent coefficients of the principal components of the hand pose given to MANO.
We finally have $\theta_f$, which are force parameters explained in Section~\ref{sec:computed_quantities}.
We concatenate all optimization variables into $\theta = [ \theta_o, \theta_h, \theta_f ]$.

\subsection{Physical model}
\label{sec:computed_quantities}

\begin{figure}
    \centering
    \includegraphics[width=1\linewidth]{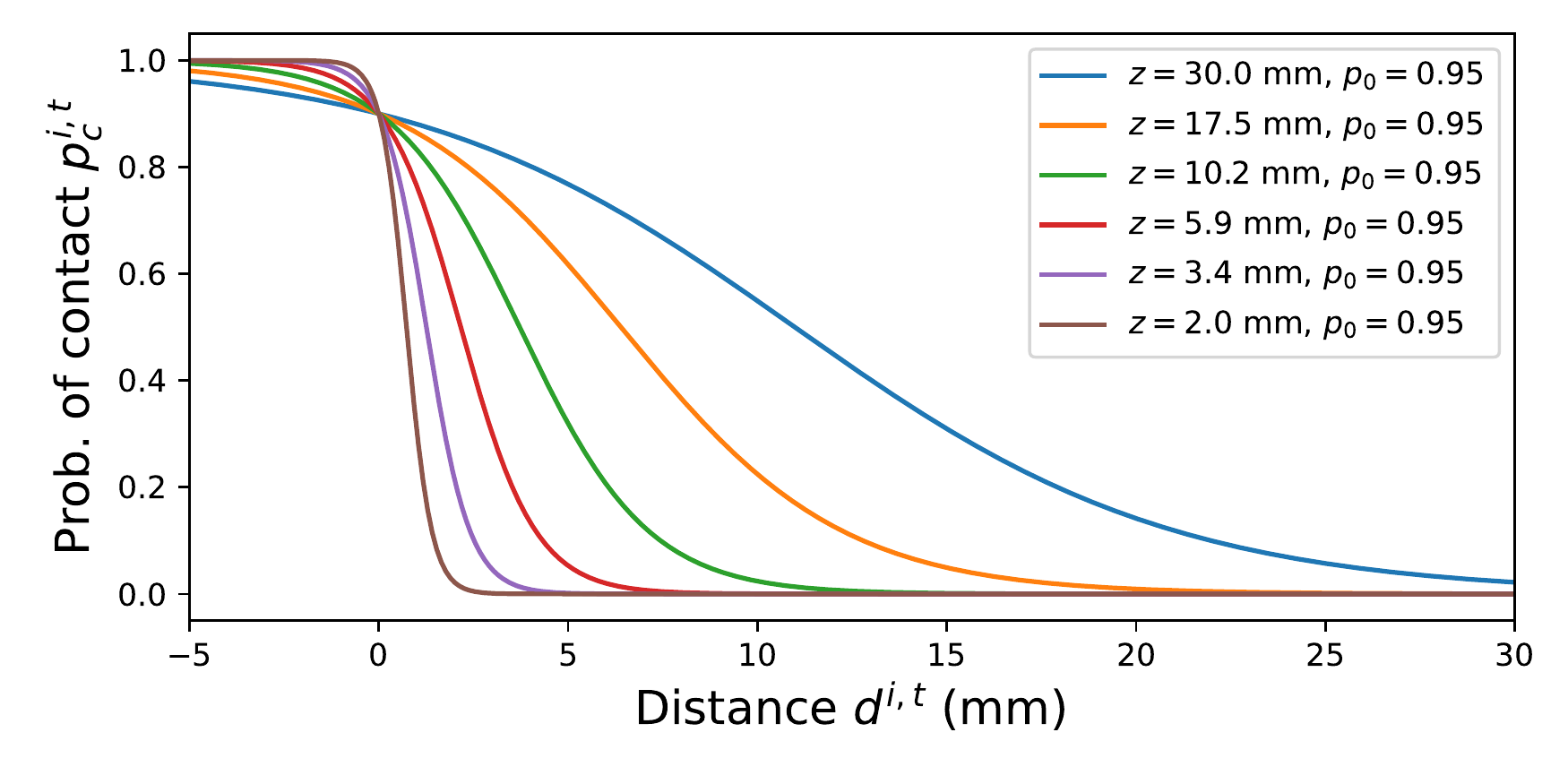}
    \caption{Probability of contact as a function of distance to the closest point on the hand, while varying the hyper-parameter $z$.}
    \label{fig:d2prob_contact}
\end{figure}

Using $\theta_h$ and $\theta_o$, we first compute the mesh vertex positions and mesh vertex normal vectors for both the hand and the object.
For simplicity, we will only denote the attributes of the object.
We denote each vertex of the object $i$ at each frame of the video, $t$, with position $\textbf{v}^{i,t}$ and normal $\textbf{n}^{i,t}$. For each vertex, we calculate the signed distance,
$d^{i,t}$, 
of the vertex to the closest point on the hand.
We compute a differentiable contact probability for each vertex with $p_c^{i,t} = \sigma\left(-\frac{6}{z}d^{i,t}-\ln\left(\frac{1}{p_0}-1\right)\right)$, where $z$ and $p_0$ are hyperparameters that determine the width of the 
function and the probability of contact for zero distance, respectively.
Figure~\ref{fig:d2prob_contact} shows how the probability of 
contact may vary as a function of distance to the closest hand vertex.

We additionally compute, for each vertex, a normal force, $\textbf{f}_n^{i,t}$,
and a static friction force, $\textbf{f}_s^{i,t}$,
which is orthogonal to the normal force.

We decompose the computation of the normal force $\textbf{f}_n^{i,t}$ as the product of three terms: 
\begin{itemize}
    \setlength{\itemsep}{0em}
    \item an upper bound on the magnitude, $f_{n,b}^{i,t} = F_{max}p_c^{i,t}$;
    \item a force activation, $\sigma(f_{n,a}^{i,t}) \in [0,1]$, where $\sigma(\cdot)$ is the sigmoid function;
    \item and the direction of the force, $-\textbf{n}^{i,t}$.
\end{itemize}
This construction constrains the direction of the normal force to always point into the object, and constrains its magnitude by the probability of contact.

Similarly, we decompose the static frictional force $\textbf{f}_s^{i,t}$ as the product of three terms:
\newcommand{\vectorproj}[2][]{\textit{proj}_{\vect{#1}}\vect{#2}}
\newcommand{\vect}{\mathbf}
\begin{itemize}
    \setlength{\itemsep}{0em}
    \item an upper bound on the magnitude, $ f_{s, b}^{i,t} = \mu_s\Vert\textbf{f}_{n}^{i,t}\Vert $, where $\mu_s$ is the coefficient of static friction;
    \item a force activation, $\tanh(\Vert \textbf{f}_{s,p}^{i,t}\Vert) \in [0,1]$;
    \item and the direction of the force, $\textbf{f}_{s,p}^{i,t}/\Vert \textbf{f}_{s,p}^{i,t}\Vert$, exactly orthogonal to the normal force.
\end{itemize}
$\textbf{f}_{s,p}^{i,t} $ is the force parameter $\textbf{f}_{s,a}^{i,t}$ projected onto the plane orthogonal to $\textbf{f}_n^{i,t}$.

We choose these decompositions specifically because they enable us to learn the forces while requiring the forces to be physically plausible.
And, importantly, this decomposition makes the entire computation of forces end-to-end differentiable, as the gradients of the forces can be backpropagated through the contact probabilities in order to learn the object and hand pose parameters $\theta_o$ and $\theta_h$.
This enables, for example, the optimizer to move the fingers of the hand closer to the object when needed in order to prevent the object from falling by increasing forces.

Learning separate force parameters for each vertex and frame, $\theta_f = \{ f_{n,a}^{i,t}, \textbf{f}_{s,a}^{i,t}\ \forall i,t\}$, would not scale well for longer videos or high-resolution meshes.
Additionally, these quantities are relatively continuous in space and time due to our continuous and differentiable contact model.
We fix the scalability issue and implicitly add a smoothness constraint by estimating these quantities using a neural network as a smooth function approximator over space and time with $\left[ f_{n,a}^{i,t}, \textbf{f}_{s,a}^{i,t} \right] = \textbf{f}_{n,a,s,a}(\textbf{v}^{i,t}, t; \theta_f)$. 
The size of the network's parameters, $\theta_f$, are independent of mesh density and video length.
In practice, we used a 6-layer feedforward network with exponential linear unit (ELU) activations.

We represent the static friction parameter, $\textbf{f}_{s,a}^{i,t}$, as a 3D vector projected onto a plane, rather than in the 2D plane itself, due to the implications of the hairy ball theorem around the mesh, which states that 2D vector fields around the 3D mesh cannot be smooth without having zeros.

The net learned force on the object per frame is found with $\textbf{f}_{net}^t = m \bm{g} +  \frac{1}{N_v}\sum_i^{N_v}  \textbf{f}_n^{i,t}+\textbf{f}_s^{i,t}$, where $m$ is the mass of the object and $\bm{g} = -9.8 \jhat \, \si{\meter \per \square \second}$.
We average the per-vertex forces, rather than sum them, to make the model agnostic to the number of vertices sampled from the mesh, $N_v$.

The finite difference velocity and acceleration are calculated with $\dot{\textbf{x}}^t = \textbf{x}^t-\textbf{x}^{t-1}$ and $\ddot{\textbf{x}}^t = \dot{\textbf{x}}^t-\dot{\textbf{x}}^{t-1}$, where $\textbf{x}^t$ is the translation vector from $\theta_o$ at frame $t$~\cite{hampaliHOnnotateMethod3D2020}. 
The finite difference net force is calculated with $\textbf{f}_{{net},{fd}}^t = m\ddot{\textbf{x}}^t$.

\subsection{Energy function}
\label{sec:energy_function}
We define the full energy function as
\begin{align}
\label{eqn:full_energy}
    E(\theta) =\ 
    &\gamma_{phy}E_{physics}(\theta) +
    \gamma_{fr}E_{force\_reg}(\theta)\ + \\
    &\gamma_{pen}E_{pen}(\theta)+
    \gamma_{d}E_{dev}(\theta) + \gamma_{s}E_{smooth}(\theta) \nonumber
\end{align}
where the optimal solution is that which minimizes the total energy,
\begin{equation}
    \theta^{*} = \argmin_{\theta} E(\theta)
\end{equation}
\noindent This problem is highly non-convex, so we use the Adam optimizer to solve this problem locally with gradient descent.

\paragraph{Physics term.}
The physics loss term matches the net learned force to the observed motion of the object by using the finite difference net force.
\begin{equation}
E_{physics}(\theta) = \sum_t
\Vert \textbf{f}_{net, fd}^t - \textbf{f}_{net}^t\Vert_2^2
\end{equation}
We tested an analogous term for torque, but it did not change the resulting forces and was removed for brevity.

\paragraph{Force minimization term.}
The problem of estimating per-vertex forces using the loss term  above alone is ill-posed, as many sets of possible forces can achieve the same net force when summed.
To combat this, we impose a force regularization term that prefers per-vertex forces with a low squared-$L^2$ norm.
This incentivizes solutions with less total force rather than more total force, and many small forces rather than a few big forces.
This is especially useful when there are multiple contact points, as is usually the case in hand-object interaction.
\begin{equation}
E_{force\_reg}(\theta) = \sum_{i,t} 
\Vert\textbf{f}_n^{i,t}\Vert_2^2
+\Vert\textbf{f}_s^{i,t}\Vert_2^2 
\end{equation}

\paragraph{Interpenetration term.}
Another physical prior we impose penalizes mesh interpenetrations.
We assume that the true poses should have no interpenetrations of the meshes, aside from small deformations of the hand at contact points. 
\SI{2}{\mm} of deformation in the hand is normal due to soft-tissue dynamics~\cite{gradyContactOptOptimizingContact2021}.
We enforce this by penalizing the signed distance of every vertex in the object to the closest hand vertex if it is less than $d_{pen}=2\,\si{\mm}$.
An analogous loss is added for the hand.
\begin{equation}
E_{pen}(\theta) = \sum_{i,t} \max(0, -(d^{i,t}+d_{pen}))
\end{equation}

\paragraph{Initialization deviation term.}
Because many physically plausible pose solutions to the optimization still exist and may satisfy the energy terms above, we introduce a penalty that minimizes the deviations of the estimated poses from the initial starting point. 
Specifically, we penalize deviations between the estimated vertex positions of the object, $\textbf{v}^{i,t}$, and their original positions, $\textbf{v}_o^{i,t}$.
An analogous loss is added for the hand.
\begin{equation}
E_{dev} (\theta) = 
\sum_{i,t} \Vert \textbf{v}^{i,t}-\textbf{v}_o^{i,t}\Vert_2^2
\end{equation}

\paragraph{Smoothness term.}
Since the hand and object pose come from a temporal sequence, we added a smoothing term that minimizes the magnitude of the finite difference acceleration.
An analogous loss is added for the hand.
\begin{equation}
E_{smooth} (\theta) = \sum_t \Vert \ddot{\textbf{x}}^t \Vert_2^2
\end{equation}

\begin{figure}
    \centering
    \includegraphics[width=1.0\linewidth]{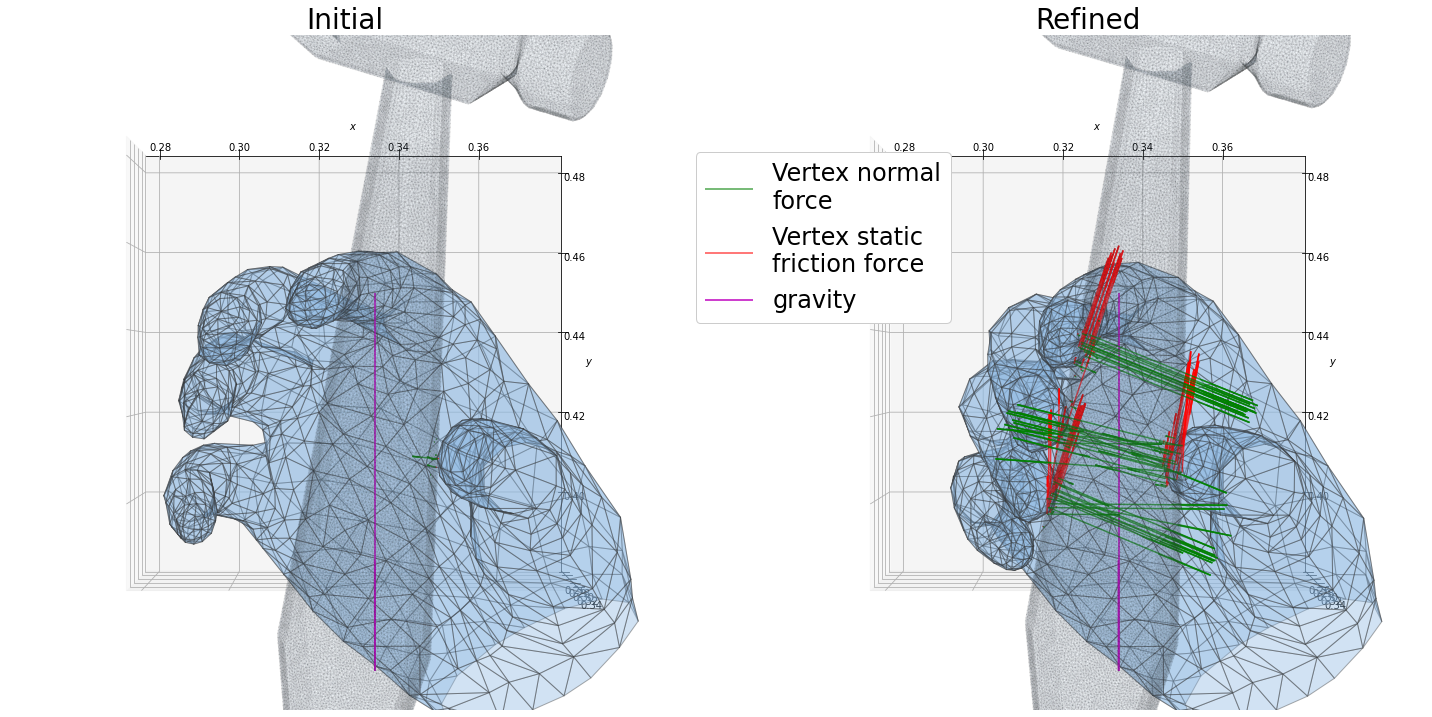}
    \caption{The hand, object, and per-vertex object normal and static frictional forces, before and after our refinement for one frame of one recording.
    Gravity is in the $-y$ direction. 
    The optimization learns to bring the hand fingers closer to the object in order to have high normal forces pointing into the object, and thus allowing high static frictional forces pointing upward to counteract gravity.}
    \label{fig:scene}
\end{figure}

\section{Evaluation}
\label{sec:eval}

We evaluated our system on ContactPose~\cite{brahmbhattContactPoseDatasetGrasps2020}, a dataset of subjects manipulating various objects in a static grasp with ground-truth hand and object pose and object contact maps.
In this section, we show that our proposed procedure can refine a noisy estimate of pose (commonly given by pose estimation systems) to closer match the the ground truth pose and contact.

We initialized our system with hand and object DoF that match the ContactPose ground truth meshes and then introduced noise, emulating a noisy pose estimation system.
Specifically, we multiplied the MANO finger pose parameters with noise from $\Delta \theta_h \sim \mathcal{N}(1, 0.1)$.
We then computed our predicted object contact map by using the probability of contact, $p_c^{i,t}$, for each vertex in the object mesh as described in Section~\ref{sec:computed_quantities}.
After performing energy minimization with the model from Section~\ref{sec:energy_function}, we had initial and refined poses, and their corresponding contact maps.
We performed this procedure on each of the one-handed object recordings included in the ContactPose dataset.

We performed our optimization with a batch size of $40$ frames, defining an epoch to be one run through the entire video.
Each batch, we randomly sampled $5000$ vertices from the hand and object mesh to be used for all computed quantities.
During optimization, we varied $z$ from $30 \,\si{\mm}$ to $2\,\si{\mm}$ logarithmically each epoch to simulate going from a very gradual, smooth description of contact to more sudden, realistic one as shown in Figure~\ref{fig:d2prob_contact}.
We ran $300$ epochs with $F_{max}=5 \,\si{\N}, 
    \gamma_{phy}=\num{5e2},
    \gamma_{fr}=\num{3e-1}$,
    $\gamma_{pen}=\num{50e6},
    \gamma_{d}= \num{2e6},
    \gamma_{s}=\num{1e5}$.
Since ContactPose assumes a rigid grasp of the object, we shared finger pose across all frames.

We compared our hand pose to the ground truth hand pose with the Mean Per-Joint Position Error (MPJPE)~\cite{gradyContactOptOptimizingContact2021}, which is the average $L^2$ norm of the error for each joint on the hand.
After thresholding the ground truth contact maps at $0.4$ \cite{brahmbhattContactPoseDatasetGrasps2020}, we compared our predicted contact map to this binary ground truth contact map by calculating the area under the precision-recall curve (PR-AUC) and the receiver operating characteristic curve (ROC-AUC).
We chose these metrics since they are invariant to monotonic transformations of the predicted contact maps, and are independent of the rectifier used to obtain predicted contact probabilities.

\begin{table}[]
    \centering
    \begin{tabular}{|c || c | c | c | c |}
     \hline
       & \makecell{MPJPE \\ (mm)} & \makecell{PR-AUC\\(\%)} & \makecell{ROC-AUC\\(\%)} \\
     \hline
     Initial & 8.95 & 71.84 & 87.58\\
     \hline
     Refined        & \textbf{4.30} & \textbf{79.76} & \textbf{91.22}\\
     \hline
    \end{tabular}
    \vspace{10pt}
    \caption{Results for the pose (MPJPE) and contact metrics (PR-AUC and ROC-AUC), before and after applying our method.}
    \label{tab:eval_table}
\end{table}

\begin{figure}
    \centering
    \includegraphics[width=0.9\linewidth]{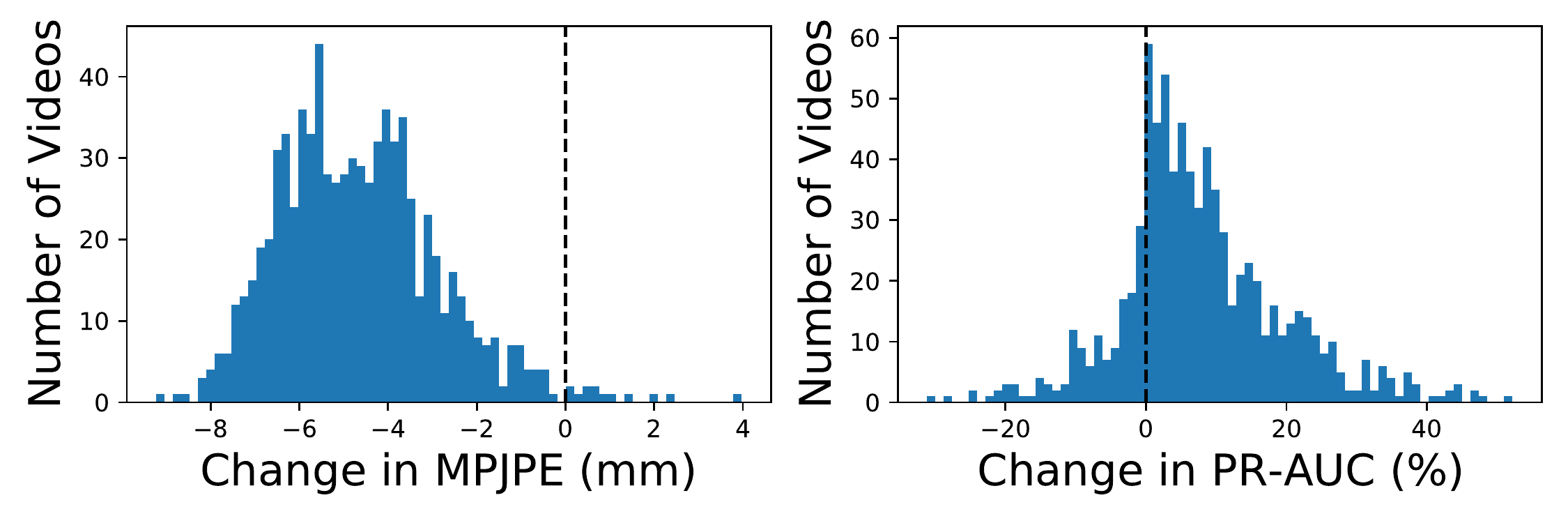}
    \caption{Distribution of the change in MPJPE and PR-AUC after refinement across all participants and objects.}
    \label{fig:eval_hist}
\end{figure}

Table~\ref{tab:eval_table} presents our results before and after our refinement method, showing all quantities improved.
The change in MPJPE and PR-AUC after refinement for all the recordings is reported in the histogram in Figure~\ref{fig:eval_hist}.
These results show that our physics-based refinement improved pose estimation (MPJPE) in $\SI{98.38}{\percent}$ of videos and contact estimation (PR-AUC and ROC-AUC) in $\SI{77.75}{\percent}$ of videos.
Overall, our system improved MPJPE by an average of $\SI{4.65}{\mm}$, PR-AUC by $\SI{7.93}{\percent}$, and ROC-AUC by $\SI{3.64}{\percent}$.
One example of the results of the optimization is shown in Figure~\ref{fig:scene}.

We achieve these gains \emph{despite not having any terms that use observed data from the video} in our optimization.
Adding such terms may further aid refinement.

The refined poses are physically plausible, which is a vital quality to have in any application that requires precise poses in hand-object interaction.
Moreover, our system provides forces that match the motion of the object, which can be used in any application that needs the physics of a scene.

\section{Conclusion}

We propose a procedure that uses temporal information in hand-object interaction to refine hand and object pose.
We demonstrate that this method improves pose and contact estimation on the ContactPose dataset given noisy estimates of the pose.
Additionally, our method is modular, since the initial hand and object pose can come from any external pose estimation system.
This generality lets it be applied to any object manipulation setting wherein hand and object pose must be accurate and physically plausible -- namely, AR and VR applications.
In the future we aim to extend our procedure to work with dynamic grasps and more entities, enabling pose refinement in more complex scenarios.

{\small
\bibliographystyle{ieee_fullname}
\bibliography{main}
}

\end{document}